\pdfoutput=1
\documentclass[11pt]{article}

\usepackage[a4paper,margin=2.2cm]{geometry}
\usepackage{natbib}
\usepackage{times}
\twocolumn

\usepackage{latexsym}
\usepackage{microtype}
\usepackage{inconsolata}
\usepackage{amsmath,amssymb}
\usepackage{booktabs}
\usepackage{multirow}
\usepackage{graphicx}
\usepackage{tikz}
\usepackage{url}
\usepackage{xspace}
\usepackage{enumitem}
\usepackage{array}
\usepackage{tabularx}
\usepackage{placeins}
\usepackage{float}
\newcommand{\dQ}{\ensuremath{\Delta Q}\xspace}
\newcommand{\Cpol}{\ensuremath{C_{\mathrm{policy}}}\xspace}
\newcommand{\Lon}{\ensuremath{L_{\mathrm{online}}}\xspace}

\title{Beyond the Query:\\
Do Retrieval Signals Improve Adaptive Multimodal RAG Routing?}

\author{%
\begin{tabular}{@{}ccc@{}}
\parbox[t]{0.30\textwidth}{\centering
\textbf{Qiaomu Li}\\
\small College of Computing and Software Engineering\\
Kennesaw State University\\
Marietta, GA 30060, USA\\
qli12@students.kennesaw.edu}
&
\parbox[t]{0.30\textwidth}{\centering
\textbf{Qiuyuan Zhang}\\
\small College of Computing and Software Engineering\\
Kennesaw State University\\
Marietta, GA 30060, USA\\
qzhang12@students.kennesaw.edu}
&
\parbox[t]{0.30\textwidth}{\centering
\textbf{Nong Ming}\\
\small College of Computing and Software Engineering\\
Kennesaw State University\\
Marietta, GA 30060, USA\\
nming@students.kennesaw.edu}
\end{tabular}}
\date{}

\begin{document}
\maketitle

\begin{abstract}
Adaptive RAG often uses retrieval-time signals to decide whether another retrieval, reranking, or multimodal step should run. We ask whether these signals add routing value once the query itself is already known. Across document, audio, and video RAG, we compare matched query-only and query+retrieval routers while holding the optional actions, router family, training procedure, and evaluation fixed. On the held-out final evaluation, adding the tested retrieval signals does not produce a reliable routing improvement over the query-only baseline. Some retrieval signals are associated with whether a later step will help, but that predictability does not consistently lead to better RUN/SKIP decisions. The main lesson is therefore methodological: retrieval-state features should not be credited with routing value unless they improve over a matched query-only control. Our results do not show that routing or retrieval state is generally useless; they show that the incremental value of retrieval signals must be demonstrated rather than assumed.
\end{abstract}

\section{Introduction}

Retrieval-augmented generation (RAG) gives a language model external evidence before it answers a query \citep{lewis2020rag}. A standard pipeline retrieves evidence, may rerank it, builds a context, and generates an answer. Adaptive RAG adds a decision: should the system run another retrieval, reranking, or multimodal step, or should it stop and answer? \citep{jiang2023flare,asai2024selfrag,jeong2024adaptiverag,su2024dragin,tang2025mbarag,ding2026ragonadiet,wang2026xrouter,guo2026routerag}

Many adaptive systems use information produced by the current retrieval, such as retrieval scores, score gaps, or statistics of the retrieved evidence. The intuition is that these signals reveal whether more computation will help. But the retrieval state is itself produced from the query. A difficult query may both create unusual retrieval scores and be more likely to benefit from another RAG step. Retrieval signals can therefore correlate with future step benefit without adding useful information beyond what the query already reveals.

We test that distinction directly:
\begin{quote}
\textbf{After an initial retrieval, do the available retrieval signals improve routing beyond what the query already tells us?}
\end{quote}

Let $q$ denote the user query and let $R(q)$ denote the retrieval scores and evidence statistics produced after processing that query. Our primary comparison is
\[
q
\qquad\text{versus}\qquad
[q,R(q)].
\]
The question is not whether retrieval can be interpreted without a query. It is whether observing the query-conditioned state $R(q)$ improves the next RUN/SKIP decision after $q$ is already known.

We compare two matched routers. One sees query features only. The other sees the same query features plus $R(q)$. The optional actions, router family, training procedure, and evaluation are otherwise held fixed. This isolates the operational routing value of adding the tested retrieval signals within our evaluated setup.

The development result is positive, but the advantage does not hold clearly on the held-out final evaluation. Adding retrieval information improves the routing score by $+0.01621$ on development data. On the 720-query final set, the gain falls to $+0.00288$, and the 95\% confidence interval includes zero. Several stronger router variants also provide only small development-time improvements. These results do not prove that no future router or state representation could work. They show that the tested retrieval inputs do not provide a reliable final advantage over the query-only control.

We then inspect the optional steps to understand this result. The document experiments expose several different regimes: DOC\_CLIP has useful but weakly predictable cases; Tile D2 can remove useful context; and Whole-page D2 preserves context but still has rare, hard-to-predict benefit. Audio and Video provide the clearest contrast: predicted and observed step benefit are positively correlated, yet query+retrieval routing does not establish an advantage over query-only routing.

\paragraph{Main empirical claim.}
\textbf{In our tested multimodal RAG setting, adding the evaluated retrieval signals does not yield a reliable routing advantage beyond query features on the held-out final evaluation.} This is not a claim that routing is useless or that retrieval state contains no useful information in principle.

The academic significance is methodological. Best-case action gain, prediction correlation, and final routing quality answer different questions. To attribute routing value specifically to retrieval state, adaptive-RAG studies need a matched query-only control. Our experiments show why: a retrieval signal can be predictive of future step benefit without improving the final routing decision once query information is already available.

Our contributions are:
\begin{enumerate}[leftmargin=*,nosep]
\item a paired before/after evaluation of optional document, audio, and video RAG steps, measuring the observed answer-quality change after each step;
\item a controlled $q$ versus $[q,R(q)]$ ablation that measures the incremental routing value of the tested query-conditioned retrieval signals on a held-out final evaluation; and
\item step-level diagnostics showing how context loss, rare benefit, weak pre-action signal, failure to improve RUN/SKIP decisions, and limited serving-time leverage can constrain adaptive gains.
\end{enumerate}

\begin{figure}[t]
\centering
\fbox{\begin{minipage}{0.92\columnwidth}
\footnotesize
\textbf{Query-conditioned retrieval state}\\[1mm]
User query $q$ $\rightarrow$ initial retrieval $\rightarrow$ retrieval state $R(q)$\\[1mm]
\textbf{Matched routing test}\\
Router A: $q$ only\\
Router B: $[q,R(q)]$\\
Same optional steps, router family, training procedure, and evaluation.\\[1mm]
\textbf{Question:} Does observing $R(q)$ improve the final RUN/SKIP decision after $q$ is already known?
\end{minipage}}
\caption{The paper's primary input ablation. Retrieval state is query-conditioned, so the relevant control is query-only versus query+retrieval, not retrieval-only.}
\label{fig:audit}
\end{figure}

\section{Related Work}

\paragraph{Adaptive RAG.}
Adaptive RAG methods decide when to retrieve, what to retrieve, or how much retrieval to use \citep{jiang2023flare,asai2024selfrag,jeong2024adaptiverag,su2024dragin,guo2025dior}. Other work explicitly considers retrieval cost or chooses among resource levels \citep{tang2025mbarag,ding2026ragonadiet,wang2026xrouter,guo2026routerag}. Several successful systems use signals available after an initial inference or retrieval stage. SAGE uses features from an initial retrieval to choose retrieval depth, while X-Router uses lightweight probes to choose between retrieval and reasoning strategies \citep{raza2026sage,wang2026xrouter}. SPARKLE learns a separate retrieval policy and also reports positive adaptive-routing results \citep{fang2026sparkle}. These systems motivate a narrower attribution question: because retrieval-time signals are generated from the query, how much routing value do they add after query information is already available?

\paragraph{Best-case value, prediction, and learned routing.}
Adaptive Re-Ranking studies when expensive reranking should run and reports a gap between best-case routing opportunity and what a learned policy captures \citep{genc2026adaptivereranking}. A budget-aware Active-RAG study separately examines retrieval benefit, routing quality, harmful retrieval, budget behavior, and trigger-side cost \citep{qian2026activeeval}. Another routing study reports settings in which a strong fixed policy remains competitive \citep{nunepalli2026slo}. We therefore do not claim that oracle analysis, harm checks, prediction--decision gaps, or cost-aware evaluation are individually new. Our main comparison instead asks an attribution question that these quantities do not answer by themselves: once the query is known, do the query-conditioned retrieval signals add measurable routing value?

\paragraph{Relevant evidence is not always useful evidence.}
R$^3$AG distinguishes evidence that looks relevant from evidence that actually helps the generator answer correctly \citep{zhao2026r3ag}. Recent selective-modality work similarly shows that a modality can be relevant even when paying to use that modality is unnecessary for the final answer \citep{li2026modality}. These studies motivate our use of downstream answer change rather than retrieval relevance as the target.

\paragraph{Multimodal RAG.}
ColPali retrieves visually rich document pages directly from page images \citep{faysse2025colpali}. LAD-RAG combines neural retrieval with a document graph \citep{sourati2026ladrag}. We do not propose a new multimodal retriever. We use multimodal RAG as a test bed because an optional step can change the representation, the amount of visual detail, or the surrounding context. These changes can make the value of the next step difficult to predict from information available beforehand.

\paragraph{Prediction and decisions are different.}
Decision-focused learning evaluates predictions partly by the decisions they support rather than only by prediction error \citep{kong2025df2}. We do not claim this general principle as new. Our contribution is a controlled multimodal RAG study that asks a more specific question: does retrieval information improve routing beyond a matched query-only baseline?

\section{What We Test}

For each query, the system reaches a point where it can either run the next optional RAG step or skip it. The router must decide before seeing what that step would produce. We therefore separate the query, the retrieval state produced from that query, and the future outcome of the optional step.

Let $Q_{\mathrm{before}}$ be answer quality before an optional step and $Q_{\mathrm{after}}$ be answer quality after it. We define the observed quality change as
\[
\dQ = Q_{\mathrm{after}}-Q_{\mathrm{before}}.
\]
A positive \dQ means that the step helped. A negative value means that it hurt. A zero value means that answer quality did not change. We obtain these labels by actually running the system before and after each optional step. The router never sees future answer quality when it makes its decision.

\subsection{Primary Test: Does Query-Conditioned Retrieval State Add Value?}

Let $q$ be the user query and let $R(q)$ be the retrieval scores and evidence statistics produced by the current retrieval. We compare two matched routers:
\begin{itemize}[leftmargin=*,nosep]
    \item \textbf{Query-only router:} receives features derived from $q$.
    \item \textbf{Query+retrieval router:} receives the same query features plus $R(q)$.
\end{itemize}

The optional RAG steps, router family, training procedure, and evaluation protocol are held fixed. Only the input features change. This is therefore an input ablation within one matched routing setup:
\[
q \qquad\text{vs.}\qquad [q,R(q)].
\]
The purpose is to measure whether observing the current retrieval state improves the routing decision \emph{after the query is already known}.

\paragraph{Why the control is query-only.}
The retrieval state is generated by applying the retrieval system to the query, so $R(q)$ is query-conditioned by construction. A retrieval-only router would answer a different deployment question. The relevant control is query-only because we want to measure what $R(q)$ adds after the query is already available.

Both routers predict the quality change from running the next step. During the controlled comparison, each optional step also has the same fixed relative evaluation cost, \Cpol, for both routers. For a cost weight $\lambda$, the router uses
\[
\hat U = \widehat{\dQ}-\lambda\Cpol.
\]
In plain terms, the router runs the step when its predicted quality gain is large enough to justify the assigned cost. The relative cost is part of the evaluation protocol; it is not measured serving latency.

We summarize each router across the same cost settings with a routing score $F_m$ for modality $m$. The incremental routing value from adding retrieval signals is
\[
G_m=F_m(r_{\mathrm{query+retrieval}})-F_m(r_{\mathrm{query}}).
\]
A positive $G_m$ means that adding $R(q)$ improved routing under the matched evaluation. We average Document, Audio, and Video equally:
\[
G_{\mathrm{macro}}=\frac{1}{3}\sum_{m\in\{D,A,V\}}G_m.
\]
This difference is the paper's primary outcome. It is an operational measure of incremental routing value under the tested representation and router family. It is not a prediction-accuracy score, wall-clock time saved, or a measure of conditional mutual information.

\paragraph{How to interpret a near-zero difference.}
A near-zero $G_m$ does not prove that the full retrieval state contains no useful information in principle. It means that the evaluated representation of $R(q)$ did not yield a reliable routing advantage beyond $q$ under the tested router family and protocol. A different state representation, model, supervision signal, or decision point could expose additional value.

\paragraph{What counts as added routing value?}
The key question is comparative: does adding $R(q)$ improve RUN/SKIP decisions over the query-only control? A positive correlation between predicted and observed step benefit is useful supporting evidence, but it is not enough by itself. The added retrieval features must improve the routing outcome.

\subsection{Supporting Checks: Why Might the Incremental Gain Be Small?}

The matched router comparison tells us whether the tested retrieval signals improve routing. It does not by itself explain why the gain is large or small. We therefore use three supporting checks. These checks are interpretive diagnostics, not a causal or additive decomposition of the main result.

\paragraph{Does the optional step have useful value to capture?}
We count how often the step improves answer quality. We also report a \emph{best-case gain}: how much quality could improve if a perfect selector somehow knew in advance exactly which queries would benefit. Formally,
\[
H_{\mathrm{ideal}}=\mathbb{E}[\max(Q_{\mathrm{base}},Q_a)-Q_{\mathrm{base}}].
\]
If this value is already tiny, no router has much opportunity.

\paragraph{Can the available inputs predict which queries benefit?}
We compare predicted and observed \dQ with Spearman rank correlation on held-out predictions. We report this for query-only and query+retrieval inputs when available. The important point is that a positive correlation in isolation does not establish incremental value beyond the query. That requires the matched routing comparison above.

\paragraph{What does the optional step itself do, and what does it cost?}
An optional step can change or remove evidence, and it can also be too cheap for selective skipping to save much time. For document actions, we therefore inspect the evidence transformation and measure warm incremental latency when valid timing data exist. These checks help interpret a weak routing result, but they are not part of the primary controlled router comparison.

\section{Experimental Setup}

\paragraph{Final checked pipeline.}
All headline results come from the final multimodal pipeline after checks for information leakage and invalid experimental branches. The generator receives document images, audio waveforms, or video frames directly. Older branches that replaced these inputs with text-only substitutes are not used in the final evidence. The router also does not receive information that would only be available after the optional step. Qwen2.5-Omni-7B generates answers from the multimodal evidence \citep{xu2025qwen}. Appendix~\ref{app:provenance} records the internal experiment identifiers and exclusions for reproducibility.

\paragraph{Benchmarks and answer quality.}
Document experiments use DocVQA \citep{mathew2021docvqa}, with an ANLS-style answer-quality score. Audio experiments use Clotho-AQA \citep{lipping2022clothoa}. Video experiments use NExT-QA \citep{xiao2021nextqa}. We use $Q$ for the answer-quality score produced by the final evaluation pipeline, and all reported $Q$ values lie between 0 and 1. For Audio and Video, the preserved experiment artifacts contain the final quality values but do not name a more specific evaluator. We therefore call those values simply ``answer quality.''

The full before/after evaluation contains 3,574 queries and 10,722 system states that we actually evaluated. The held-out final evaluation contains 720 queries, with 240 from each modality. None of these 720 queries were used to train the router or choose among router designs. The preserved artifacts support the claim that these queries were kept out of router training and model selection. They do not establish that every underlying document, audio source, or video is unique across all splits, so we do not make that stronger claim.

\paragraph{The two main routers.}
Both main routers use the same histogram gradient-boosting model and training procedure. Both predict whether the optional step will increase or decrease answer quality. The difference is their input. The query-only router sees query features. The second router sees the same query features plus retrieval and evidence statistics that already exist before the optional step runs. We call this second model the \emph{query+retrieval router}. Because the actions, model family, and training procedure are otherwise matched, this comparison directly tests the incremental routing value obtained from adding current retrieval information within our evaluated setup.

\paragraph{Statistics.}
We use paired bootstrap confidence intervals to show the uncertainty around routing-score differences. To measure how well predicted benefit tracks observed benefit, we use standard Spearman rank correlation with average ranks for ties. An earlier analysis used a different tie treatment for Audio and Video. We corrected that analysis and do not use the earlier values here.

\section{Results}

\subsection{RQ1: Do Retrieval Signals Add Routing Value Beyond the Query?}

\begin{figure}[t]
\centering
\begin{tikzpicture}[x=2.35cm,y=1.00cm]
  \draw[->] (-0.45,-1.05) -- (-0.45,3.35);
  \draw[->] (-0.45,0) -- (1.40,0);
  \foreach \y/\lab in {-1/{-0.01},0/{0.00},1/{0.01},2/{0.02},3/{0.03}} {
    \draw (-0.50,\y) -- (-0.40,\y);
    \node[anchor=east,font=\scriptsize] at (-0.53,\y) {\lab};
  }
  \node[rotate=90,font=\scriptsize] at (-1.05,1.15) {Gain from adding retrieval information};
  \node[font=\scriptsize] at (0,-0.38) {Development};
  \node[font=\scriptsize] at (1,-0.38) {Final};

  \draw[line width=0.7pt] (0,0.436) -- (0,2.960);
  \draw[line width=0.7pt] (-0.08,0.436) -- (0.08,0.436);
  \draw[line width=0.7pt] (-0.08,2.960) -- (0.08,2.960);
  \fill (0,1.621) circle (1.8pt);

  \draw[line width=0.7pt] (1,-0.726) -- (1,1.107);
  \draw[line width=0.7pt] (0.92,-0.726) -- (1.08,-0.726);
  \draw[line width=0.7pt] (0.92,1.107) -- (1.08,1.107);
  \fill (1,0.288) circle (1.8pt);
\end{tikzpicture}
\caption{How much the routing score improves when the router receives current retrieval information in addition to the query. The gain is clear on development data but much smaller on the held-out final evaluation. The final confidence interval includes zero.}
\label{fig:generalization}
\end{figure}
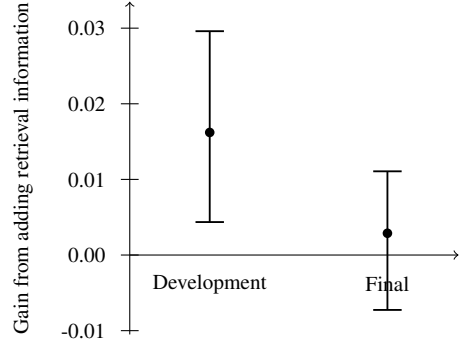

Figure~\ref{fig:generalization} shows the main result. On the development set ($N=715$), the query+retrieval router improves the routing score over the query-only router by
\[
+0.01621,\qquad 95\%\ \mathrm{CI}=[0.00436,0.02960].
\]
On the held-out final evaluation ($N=720$), the improvement falls to
\[
+0.00288,\qquad 95\%\ \mathrm{CI}=[-0.00726,0.01107].
\]
The final confidence interval includes zero. We therefore cannot conclude that adding current retrieval information gives a reliable routing improvement on the final set. The modality-level differences are $+0.00290$ for Document, $+0.00575$ for Audio, and $0$ for Video. Their equal-weight average is $+0.00288$.

This is the paper's main result. The tested retrieval signals look useful on development data, but their advantage over the query-only control becomes small and statistically inconclusive on the held-out final set. Generic prediction error does not explain the drop; it improves numerically for both routers. The finding is therefore specific to the routing decision under the evaluated representation of $R(q)$ and router family. We next test stronger routers and then inspect the optional actions for possible explanations.

\subsection{RQ2: Is the Router Simply Too Weak?}

We keep the optional RAG steps fixed and change only the router. We test five approaches. One directly predicts RUN or SKIP. One first predicts whether any gain will occur and then predicts the size of that gain. One predicts the signed quality change directly. We also test a larger 139-feature representation and a training objective that gives more weight to costly routing mistakes.

To compare these variants during development, we use the same train-only score across held-out folds and relative-cost settings. This score is called \emph{robust regret}; lower is better. Many optional steps produce exactly zero quality change, which makes direct RUN/SKIP classification difficult. In the development analysis, mistakes that skip a helpful step contribute more regret than mistakes that run a non-helpful step. Directly predicting the positive or negative quality change works better in this study. Table~\ref{tab:controller} shows the matched baseline and the strongest representative alternatives; the full set of tested variants is listed in the appendix.

\begin{table}[t]
\centering
\small
\begin{tabular}{lc}
\toprule
Router design & Robust regret $\downarrow$\\
\midrule
Query+retrieval baseline & 0.149818\\
Direct quality-change prediction & \textbf{0.148959}\\
Decision-weighted training & 0.149569\\
\bottomrule
\end{tabular}
\caption{Comparison of router designs while the optional RAG steps stay fixed. Robust regret is used only for development-time model comparison; lower is better. The best design improves over the baseline by about 0.57\%.}
\label{tab:controller}
\end{table}

The best design improves this development-time score by only about $0.57\%$ relative to the baseline. The larger feature set and decision-weighted training also change the result only slightly. This does not prove that no stronger router could ever work. It shows that, within the tested variants, the weak final advantage of $R(q)$ is not obviously explained by one simple or underpowered router. We therefore inspect the optional steps and the information they expose.

\subsection{RQ3: What Do the Document Steps Reveal About the Available Signal?}

The document experiments use three optional visual steps with internal names. We first define the steps. We then report their results separately.

\subsubsection*{The three document steps}

\paragraph{DOC\_CLIP: whole-page visual retrieval.}
The base document pipeline first uses text extracted from the document. DOC\_CLIP adds a visual retrieval stage after that text-based stage. Each document page is rendered as an image, and CLIP scores how well the page image matches the query. DOC\_CLIP can therefore retrieve a page from visual content that may not be captured well by OCR text.

\paragraph{Tile D2: stronger visual matching on page regions.}
Tile D2 is a more fine-grained visual step. It divides a page into smaller regions, or tiles, and applies a stronger multivector visual retriever to those regions. The retriever selects local regions that appear most relevant to the query. In the evaluated implementation, selected tiles can replace earlier evidence, which can remove the surrounding page context.

\paragraph{Whole-page D2: stronger visual matching while keeping page context.}
Whole-page D2, called D2$'$ in the internal experiment logs, was introduced to test the context-loss problem in Tile D2. It keeps the stronger visual matching but scores and reranks whole pages rather than returning isolated tiles. The full page therefore remains available to the generator.

\subsubsection*{Results for the three document steps}
The definitions above describe only what each step does. We now turn to the measured outcomes. Table~\ref{tab:doc} summarizes the results. The paragraphs below interpret each step separately. For these document diagnostics, $\rho$ measures predictability from the evaluated pre-action features; it should not be interpreted as incremental value beyond the query unless a matched query-only comparison is available.

\begin{table*}[t]
\centering
\scriptsize
\setlength{\tabcolsep}{3.5pt}
\begin{tabular}{lrrrrrrl}
\toprule
Step & Help & Best gain & Q+retrieval $\rho$ & Query-only $\rho$ & Extra/base & Mean $\dQ$ & Main observation\\
\midrule
DOC\_CLIP & 0.095 & 0.0792 & 0.1436 & -- & 0.244 & +0.0700 & useful cases; weak state signal\\
Tile D2 & 0.040 & -- & $-$0.0305 & -- & -- & $-$0.0141 & can remove useful context\\
Whole-page D2 & 0.0467 & 0.0368 & 0.0100 & 0.0356 & 0.0416 & +0.0203 & rare benefit; no state advantage\\
\bottomrule
\end{tabular}
\caption{Results for the three optional document steps. Help is the fraction of queries whose answer quality improves. Best gain is the improvement available to a perfect selector. Q+retrieval and Query-only $\rho$ are Spearman correlations between predicted and observed step benefit for the available pre-action models; ``--'' means that a comparable query-only value is not available for that diagnostic. Extra/base compares added median latency with the required previous stage.}
\label{tab:doc}
\end{table*}

\paragraph{DOC\_CLIP: useful cases exist, but they are only weakly predictable.}
We evaluate DOC\_CLIP on a fresh set of $N=200$ queries with no source overlap with the earlier sample. It improves 19 queries, leaves 179 unchanged, and harms 2. Its benefit is only weakly predictable from the evaluated pre-action features ($\rho=0.1436$). Helpful cases are $2.63\times$ as common in the 20\% of queries with the highest predicted benefit as in the full set, so there is some signal, but it is limited. DOC\_CLIP adds 21.76\,ms of median latency, while the required base stage takes 89.26\,ms. Thus, DOC\_CLIP has real useful cases, but the available predictive signal is weak and the maximum serving-time saving from skipping it is limited.

\paragraph{Tile D2: the action itself can make the target worse.}
Tile D2 improves only 4\% of queries and harms 7\%. Its average quality change is negative. Artifact analysis shows that selected tiles can replace useful evidence and remove broader page context. This supports a careful interpretation of weak routing results: the value being predicted depends on what the optional step actually does. If the step can damage the evidence, changing only the router may not address the underlying problem.

\paragraph{Whole-page D2: fixing context does not create an easy prediction problem.}
Whole-page D2 preserves the full page and therefore removes the main context-loss mechanism observed for Tile D2. However, it helps only 4.67\% of queries. The query+retrieval prediction has almost no rank association with benefit ($\rho=0.0100$), while the available query-only diagnostic is also very small ($\rho=0.0356$). It adds 42.86\,ms of median latency, while the required base stage takes about 1030.32\,ms. The revised action is safer, but useful cases remain rare and hard to identify before execution. Action redesign can remove one failure mode without making the future value of the step easier to predict.

\subsection{RQ4: Can a Signal Exist Without Adding Value Beyond the Query?}

Audio and Video show why the answer can be no (Table~\ref{tab:av}). These are separate step-level diagnostics, not the primary 720-query comparison in RQ1. Their purpose is to compare predictability with realized routing value for individual actions.

\begin{table*}[t]
\centering
\small
\begin{tabular}{lrrrrr}
\toprule
Modality & Help rate & Best-case gain & Query+retrieval $\rho$ & Query-only $\rho$ & Routing-score difference\\
\midrule
Audio & 0.180 & 0.130 & 0.3054 & 0.3397 & $-$0.0835\\
Video & 0.120 & 0.065 & 0.4923 & 0.4569 & $-$0.0006\\
\bottomrule
\end{tabular}
\caption{Audio and Video step-level diagnostics. The optional steps sometimes help, and their benefit can be predicted to some degree. Even so, adding retrieval information does not improve the routing score over using the query alone. Real serving-time savings remain unresolved for these modalities.}
\label{tab:av}
\end{table*}

For Audio, the optional step helps 18\% of cases, and the best-case gain is 0.130. The Spearman correlation between predicted and observed benefit is $0.3054$ for the query+retrieval router (95\% CI $[0.2183,0.3887]$) and $0.3397$ for the query-only router (CI $[0.2644,0.4140]$). In other words, both routers can rank helpful cases to some degree. Yet adding retrieval information makes the routing score 0.0835 lower in this diagnostic setting.

Video shows a similar pattern. The Spearman correlation is $0.4923$ for the query+retrieval router (CI $[0.4125,0.5664]$) and $0.4569$ for the query-only router (CI $[0.3771,0.5334]$). The first number is slightly larger, but the evidence does not establish a meaningful predictive advantage. More importantly, the routing-score difference is essentially zero ($-0.0006$).

These examples sharpen the paper's main point. A positive correlation with step benefit is not enough to establish that retrieval features add value beyond the query. The added features must improve the final RUN/SKIP routing outcome. Audio and Video show measurable predictability, but little or no realized routing gain from adding retrieval information in these step-level tests.

\section{Discussion}

\paragraph{What the main result means.}
Our study asks whether the query-conditioned retrieval state $R(q)$ improves routing after the query $q$ is already known. The matched query-only baseline is essential because query difficulty can influence both retrieval behavior and the usefulness of later computation. On the held-out final evaluation, adding the tested retrieval signals does not provide a reliable routing improvement over that baseline.

\paragraph{Predictive does not necessarily mean incrementally useful.}
Audio and Video make this distinction concrete. In both cases, predicted and observed step benefit are positively correlated, yet query+retrieval routing does not outperform query-only routing in the step-level diagnostics. A retrieval signal can therefore look predictive without showing that the retrieval state adds routing value once the query is already available. This is the paper's main methodological point.

\paragraph{Why this is not a claim that routing is unnecessary.}
A system may still benefit from routing. Our result concerns the particular retrieval features, representations, router families, and decision points we tested. A different state representation, observable signal, or decision point could perform better. The stronger-router experiments also do not prove that the retrieval state contains no additional information; they only show that the tested model changes do not recover a clear advantage from the same general signals. When query+retrieval does not beat query-only, a larger router should therefore not be the automatic next step.

\paragraph{The optional action also matters.}
The document experiments show why. Tile D2 can remove context and make answers worse. Whole-page D2 fixes that mechanism, but useful cases remain rare and both query-only and query+retrieval predictability are weak. DOC\_CLIP has useful cases, but its measured serving-time leverage is limited. These diagnostics do not causally decompose the main $q$ versus $[q,R(q)]$ result. They show that routing difficulty depends both on the information available at the decision point and on the action whose future value is being predicted.

Table~\ref{tab:audit-summary} summarizes these supporting diagnostics.

\begin{table}[H]
\centering
\scriptsize
\renewcommand{\arraystretch}{1.08}
\begin{tabularx}{\columnwidth}{>{\raggedright\arraybackslash\bfseries}p{0.29\columnwidth}>{\raggedright\arraybackslash}X}
\toprule
Observed pattern & Evidence and interpretation\\
\midrule
Action can hurt evidence & Tile D2: 4\% help, 7\% harm, negative mean $\dQ$. The action itself can limit routing quality.\\
\addlinespace
Rare, hard-to-predict benefit & Whole-page D2: 4.67\% help; Q+retrieval $\rho=0.0100$, query-only $\rho=0.0356$. Fixing context loss does not make future benefit easy to predict.\\
\addlinespace
Predictive, but no extra routing value & Audio/Video have positive query+retrieval and query-only correlations, yet query+retrieval versus query-only routing differences are $-0.0835$ and $-0.0006$. Predictability alone does not establish incremental retrieval-state value beyond the query.\\
\addlinespace
Useful, but relatively cheap & DOC\_CLIP improves 19/200 queries and adds 21.76\,ms over an 89.26\,ms base stage. Cost limits possible serving payoff; this is separate from the main input ablation.\\
\bottomrule
\end{tabularx}
\caption{Supporting diagnostics for the query-only versus query+retrieval comparison. These rows are not mutually exclusive failure classes or an additive decomposition. Audio/Video serving-time savings remain unresolved.}
\label{tab:audit-summary}
\end{table}

\paragraph{Implication for adaptive-RAG evaluation.}
Best-case gain asks whether an optional step could help with perfect knowledge. Prediction correlation asks whether predicted benefit tracks observed benefit. Final routing quality asks whether the RUN/SKIP policy actually improves. None of these alone attributes value specifically to retrieval state. When a router uses query-conditioned retrieval features, a matched query-only control provides that missing comparison. In our experiments, this control changes the interpretation: optional steps can have real utility and measurable predictability even when the tested retrieval features add no reliable routing advantage beyond the query.

\paragraph{Academic significance.}
The contribution is a boundary-condition and measurement result rather than a new universal router. It identifies a failure mode that can be hidden by predictive metrics: retrieval-state predictability need not become incremental routing value. This distinction helps determine the next research step---a larger router, a better state representation, a different action, or a different decision point.

\FloatBarrier

\section{Limitations}

The held-out final result is small and statistically inconclusive, so we do not claim that we built a better router or that retrieval-state features are generally ineffective. We tested several stronger router designs, but many other models and state representations remain possible.

Real incremental retrieval time is unresolved for Audio and Video because the stored zero timing values were placeholders rather than valid warm-serving measurements. The final 720-query set was held out from router training and model selection, but the preserved artifacts do not prove complete source-level separation across every split. We therefore make only the held-out claim supported by the evidence.

We also lack a completed positive-control experiment in which retrieval features clearly improve routing over a query-only control. The planned NFCorpus control was stopped before scientific outcomes because the required source-separated split was not feasible; Appendix~\ref{app:positive-control} gives the details.

Finally, the $q$ versus $[q,R(q)]$ framing was sharpened during analysis rather than fixed as the project's original hypothesis. We study a limited set of benchmarks, one main multimodal generator, and a finite set of optional actions and retrieval features. The comparison is an operational routing ablation, not a measurement of conditional mutual information or a proof of feature redundancy. The router also makes decisions only within the current modality rather than selecting jointly across document, audio, and video.

\section{Conclusion}

Adaptive RAG often assumes that the current retrieval state helps decide whether more computation is needed. We test that assumption with a matched query-only versus query+retrieval comparison. On the held-out final evaluation, the tested retrieval signals do not provide a reliable routing advantage over the query alone. Audio and Video also show that a signal can predict future step benefit without improving the final RUN/SKIP decision.

\textbf{The main takeaway is simple: predictive retrieval signals are not the same as incremental routing value.} Retrieval-state features should be credited with routing value only when they improve over what the query already provides. This does not mean that routing is unnecessary; other representations, actions, models, or decision points may work better. But when query+retrieval does not reliably beat query-only, simply building a larger router may not address the real bottleneck. Adaptive-RAG studies that use retrieval-state features should therefore include a matched query-only control when possible.

\appendix
\section{Evidence Checks and Experiment History}
\label{app:provenance}

\paragraph{Final evidence used in the paper.}
The final scientific results come from the multimodal pipeline internally named S1\_V2. We exclude several older experimental branches. The historical Phase-3 data did not use real multimodal generation. An older L0 proxy leaked query information. Historical V1.1 results used text-only or otherwise non-native substitutes for multimodal evidence. ActivityNet is not used as the main video action family because the required authentic-media and action conditions were not met.

\paragraph{Older diagnostics that remain valid.}
S1\_V2.1 is a valid development-set ablation, but it was not chosen as the frozen baseline. Earlier Audio and Video Spearman calculations handled tied values incorrectly by assigning ordinal ranks. The final analysis uses average ranks, which is the standard treatment. Stored zero-valued Audio and Video timing fields were placeholders, not valid latency measurements.

\section{Paired Before/After Data and Final Evaluation}

For every optional RAG step, we physically evaluate the system before and after the step. The corrected paired dataset contains 3,574 queries and 10,722 evaluated states. The development split contains 2,859 training queries and 715 evaluation queries. The held-out final evaluation contains 720 queries, with 240 from each modality. It includes 2,160 state outcomes and 1,440 before/after transitions.

For runs with two optional steps, the executed sequence is
\[
s_0=L_0,\qquad s_1=L_0+L_1,\qquad s_2=L_0+L_1+L_2,
\]
with answer-quality values $Q_0,Q_1,Q_2$. The observed changes are
\[
\dQ_0=Q_1-Q_0,\qquad \dQ_1=Q_2-Q_1.
\]
These differences are measured after physically running each step. The router is trained only on information that exists before the step it is deciding about.

\begin{table}[h]
\centering
\scriptsize
\setlength{\tabcolsep}{4pt}
\begin{tabular}{lrrrr}
\toprule
Modality & $Q_0$ & $Q_1$ & $Q_2$ & Final $G_m$\\
\midrule
Document & 0.03356 & 0.10941 & 0.31823 & +0.00290\\
Audio & 0.44583 & 0.48333 & 0.48333 & +0.00575\\
Video & 0.55000 & 0.54167 & 0.54167 & 0.00000\\
\bottomrule
\end{tabular}
\caption{Final answer-quality trajectories and the routing-score difference between the query+retrieval and query-only routers. The equal-weight average of the three routing differences is $G_{\mathrm{macro}}=0.00288$. The $Q_i$ values are descriptive; they do not imply that every optional step must improve quality monotonically.}
\end{table}

\paragraph{What the held-out split does and does not establish.}
The final 720-query set was not used to train the router or choose among router designs. The preserved artifacts do not establish that the underlying source document, audio item, or video is always different across every split and benchmark. We therefore make only the held-out claim supported by the evidence. We report no source overlap only for experiments where that check was directly verified, such as the fresh DOC\_CLIP evaluation.

\section{Additional Router Variants}

During development, we tested five router designs while keeping the optional RAG steps fixed. The designs were: direct RUN/STOP classification; a two-stage model that first predicts whether any gain occurs and then predicts its size; direct prediction of positive or negative quality change; a larger 139-feature representation; and a training objective that gives more weight to costly routing mistakes.

The RUN/STOP task is highly imbalanced. RUN cases make up 8.128\% of examples and STOP cases 91.872\%. Cases where all choices are STOP make up 87.52\%, and 81.464\% of observed quality changes are exactly zero. Errors that skip a helpful step contribute more to the development-time regret score than errors that run an unnecessary step. Directly predicting the signed quality change gives the best robust-regret result, but improves over the baseline by only about 0.57\%. Robust regret is the fixed development-only score used to compare models across held-out folds and relative-cost settings. Lower values are better.

\section{Correcting Spearman Correlation for Ties}

Audio and Video contain many repeated $\dQ$ values. An earlier analysis assigned ordinal ranks to these ties. That was not the intended Spearman calculation. The final paper uses standard average ranks:
\[
\begin{aligned}
\rho_{\mathrm{Audio,state}} &= 0.30545, &
\rho_{\mathrm{Audio,query}} &= 0.33967,\\
\rho_{\mathrm{Video,state}} &= 0.49232, &
\rho_{\mathrm{Video,query}} &= 0.45687.
\end{aligned}
\]
The earlier ordinal-rank values are not used in any final table or claim.

\section{Relative Evaluation Cost and Real Serving Time}

The controlled router comparison uses the same fixed relative cost $\Cpol$ for both routers. This number is part of the evaluation protocol. It lets us compare quality and assigned action cost under exactly the same settings.

The real-time question is different. For production value, we care about how much warm online latency, \Lon, can actually be avoided when a step is skipped. We therefore keep measured serving time separate from the relative cost used by the routing evaluation.

The final Audio and Video retrieval handoff stored timing fields as \texttt{0.0} placeholders. These are not measurements from a warm timing run. We do not interpret them as zero retrieval cost. We also do not replace them with generation latency or infer missing values from model complexity. Therefore, real incremental retrieval time remains unresolved for Audio A0/A1/A2 and Video V0/V1/V2.

\section{Planned Positive Control That Could Not Be Completed}
\label{app:positive-control}

We froze a separate positive-control protocol before observing scientific outcomes. It used BEIR/\allowbreak NFCorpus, TF-IDF top-20 retrieval, optional CrossEncoder L4 reranking over the same candidates, and a planned split of 150 analysis and 150 confirmation queries. A capacity check found 323 test queries with relevance judgments, but their source-overlap graph had connected-component sizes
\[
[318,1,1,1,1,1].
\]
After the first 150 analysis queries were selected, only 12 source-separated confirmation candidates remained, far short of the required 150. We therefore stopped before running retrieval, reranking, quality, $\dQ$, best-case-selection, or routing analyses. The failed split is a methodological non-result and provides neither positive nor negative evidence about routing performance.

\end{document}